\documentclass[sigconf,natbib=true]{acmart}
\usepackage{url}
\usepackage{booktabs}
\usepackage{graphicx}
\usepackage{float}
\usepackage{multirow}
\usepackage{xcolor}
\usepackage{colortbl}
\definecolor{lightblue}{rgb}{0.93,0.95,1.0}
\definecolor{lightgreen}{rgb}{0.93,1.0,0.95}

\newcommand{\done}{{}}

\AtBeginDocument{%
  }

\setcopyright{acmlicensed}
\copyrightyear{2024}
\acmYear{2024}
\acmDOI{XXXXXXX.XXXXXXX}

\acmConference[GenAIRecP @ KDD'24]{Workshop on Generative AI in Recommendation and Personalization at KDD '24}{August 25--29,
  2024}{Barcelona, Spain}
\acmISBN{978-1-4503-XXXX-X/18/06}

\begin{document}

\title{PERSOMA: PERsonalized SOft ProMpt Adapter Architecture for Personalized Language Prompting}

\author{Liam Hebert}
\authornote{Work done while an intern at Google Research.}
\email{liam.hebert@uwaterloo.ca}
\orcid{0000-0002-8990-1520}
\affiliation{%
  \institution{Google Research}
  \city{Mountain View}
  \state{California}
  \country{USA}
}

\author{Krishna Sayana}
\email{ksayana@google.com}
\affiliation{%
  \institution{Google Research}
  \city{Mountain View}
  \state{California}
  \country{USA}
}

\author{Ambarish Jash}
\email{ajash@google.com}
\affiliation{%
  \institution{Google Research}
  \city{Mountain View}
  \state{California}
  \country{USA}
}

\author{Alexandros Karatzoglou}
\affiliation{%
  \institution{Google Research}
  \city{Mountain View}
  \state{California}
  \country{USA}
}

\author{Sukhdeep Sodhi}
\affiliation{%
  \institution{Google Research}
  \city{Mountain View}
  \state{California}
  \country{USA}
}

\author{Sumanth Doddapaneni}
\authornotemark[1]
\affiliation{%
  \institution{Google Research}
  \city{Mountain View}
  \state{California}
  \country{USA}
}

\author{Yanli Cai}
\affiliation{%
  \institution{Google Research}
  \city{Mountain View}
  \state{California}
  \country{USA}
}

\author{Dima Kuzmin}
\affiliation{%
  \institution{Google Research}
  \city{Mountain View}
  \state{California}
  \country{USA}
}

\renewcommand{\shortauthors}{Hebert et al.}

\begin{abstract}
\done
Understanding the nuances of a user's extensive interaction history is key to building accurate and personalized natural language systems that can adapt to evolving user preferences. To address this, we introduce PERSOMA, \textbf{Per}sonalized \textbf{So}ft Pro\textbf{m}pt \textbf{A}dapter architecture. Unlike previous personalized prompting methods for large language models, PERSOMA offers a novel approach to efficiently capture user history. It achieves this by resampling and compressing interactions as free form text into expressive soft prompt embeddings, building upon recent research utilizing embedding representations as input for LLMs. We rigorously validate our approach by evaluating various adapter architectures, first-stage sampling strategies, parameter-efficient tuning techniques like LoRA, and other personalization methods. Our results demonstrate PERSOMA's superior ability to handle large and complex user histories compared to existing embedding-based and text-prompt-based techniques.

\end{abstract}

\ccsdesc[500]{Computing methodologies~Natural language processing}
\ccsdesc[300]{Computing methodologies~Information extraction}
\ccsdesc[300]{Computing methodologies~Natural language generation}

\keywords{Personalization, Natural Language Processing, User Understanding, Soft Prompting, Large Language Models}

\received{31st January 2024}
\received[revised]{28 May 2024}
\received[accepted]{28 June 2024}

\maketitle

\section{Introduction}
\done
Personalized systems have become increasingly essential across various applications in today's connected digital landscape. These systems leverage insights from past interactions to tailor experiences to each user's unique preferences and needs. Personalized systems are found in diverse domains, from content recommendations like music \citep{millecamp2018controlling} and movies \citep{gomez2015netflix}, to personalized medicine  \citep{schork2019artificial,johnson2021precision}  and customized educational learning pathways \citep{chen2020artificial, pratama2023revolutionizing}.

Amid these advancements in personalization, large language models (LLMs) applications have emerged in many domains. When trained on an extensive training corpus of natural language tokens and billions of parameters, large language models have displayed exceptional emergent capabilities to tackle multiple tasks without re-training. In other words, LLMs have emerged as a "one-size-fits-all" solution to many tasks in various domains. To achieve this goal, LLMs are often trained to reproduce text from large generalist and task-agnostic datasets. However, this emphasis on generalist capabilities can limit the effectiveness of LLMs in personalizing outputs towards the needs and characteristics of specific users.

Research to bring personalization to LLMs has primarily explored the development of sophisticated text prompt templates, categorized into simple personalized text prompts, retrieval-augmented prompts, and profile-augmented prompts. Given a list of natural language descriptions of prior actions and a desired task prompt, the simple approach appends the entire set of historical descriptions to the task prompt, leveraging the LLM's inherent in-context learning abilities to personalize the output. However, given the potential for extensive user histories which may exceed the LLM's context window, some studies have proposed retrieval methods to distill this history into the most relevant segments for personalization. Other approaches have also explored using an LLM to synthesize the user's history into brief natural language profiles of user preferences before concatenating them to the desired task prompt. 

However, text-prompting approaches to personalization rely on representing the user's history in a lengthy natural language paragraph. Beyond constraints with limited context lengths, several recent works have also found that LLMs often forget information when faced with long prompts. In \citet{liu2024lost}, the authors observed that LLMs tend to leave out information in the middle of the prompt. This problem is further compounded with findings by \citet{shi2023large}, which find that LLMs are also often distracted by irrelevant context when faced with large prompts. Together, these findings raise the fundamental question of this research: \textit{Is natural language prompting an effective strategy for personalization at scale?} 

Recent research \cite{doddapaneni2024user} has explored using user embeddings as prompts to enhance the personalization capabilities of Large Language Models (LLMs). Building on this approach, we introduce PERSOMA: the \textbf{Per}sonalized \textbf{So}ft Pro\textbf{m}pt \textbf{A}dapter Architecture. PERSOMA goes beyond traditional text-based personalization by leveraging parameter-efficient finetuning and history resampling techniques. Instead of relying solely on text, PERSOMA utilizes a soft prompt adapter to condense a user's historical interactions into a compact set of expressive prompt embeddings. These embeddings are further compressed using a perceiver and mapped to the LLM's vocabulary space, ensuring prompt efficiency. This innovative approach allows PERSOMA to steer the output of a frozen LLM toward user preferences without sacrificing the model's capacity for in-context learning or further finetuning. By training the soft prompt adapter on specific tasks, PERSOMA guarantees that the prompt embeddings are contextually relevant, incorporating the necessary information from the user's history to optimize for the desired task.

Our empirical studies, utilizing the PaLM 2 model, demonstrate that PERSOMA effectively harnesses extensive user history for personalization while maintaining computational efficiency. On the MovieLens user preferences dataset \cite{doddapaneni2024user}, PERSOMA outperforms previous embedding-based techniques by 0.18 in F1 score and matches the performance of a fully finetuned text prompting baseline, all while using significantly less computational resources. We investigate PERSOMA's capabilities by comprehensively evaluating various history sampling strategies, input formats, and parameter-efficient training techniques.

\section{Related Work}
Various text-based prompting methods have been proposed to achieve personalization in language models. These approaches range from concatenating historical interactions into text prompts \cite{geng2022recommendation} to using first-stage retrievers for sampling history items \cite{lamp} and prompt rewriting to summarize critical information \cite{li2023automatic}. However, text-based prompting suffers from computational constraints due to large inputs and finetuning requirements.

Soft prompting, a parameter-efficient technique introduced by \cite{lester2021power}, addresses these challenges by adapting frozen language models to specific tasks. This approach has been extended to various domains, including multilingual tasks \cite{zhao2021discrete}, transfer learning \cite{vu2022spot}, and vision models \cite{bulat2022language}. Recently, \cite{mu2023learning} proposed using the language model itself to create more expressive task soft prompt tokens.

Soft prompting has also been explored for personalization. Methods like \cite{useradapter} and \cite{useridentifier} create trainable user-specific tokens, while \cite{wu2024personalized} uses MLP networks to create personalized soft prompts from tabular user data. UEM \cite{doddapaneni2024user} explores using dense user-item embeddings as personalized soft prompts. Recent work has also explored adapting other parameter-efficient techniques for personalization. \cite{fu2024exploring} explore injecting adapter layers in the transformer stack to adapt models towards a specific topic domain. 

Building upon UEM and previous work, our research introduces novel methods for resampling, diverse encoder architectures, parameter-efficient training, and comprehensive evaluation of various history sampling techniques. This expands the capabilities and understanding of soft prompting for personalization.

\section{Method}
\label{sec:approach}
\subsection{PERSOMA Architecture}
\done
\begin{figure*}
    \centering
    \includegraphics[width=0.8\linewidth]{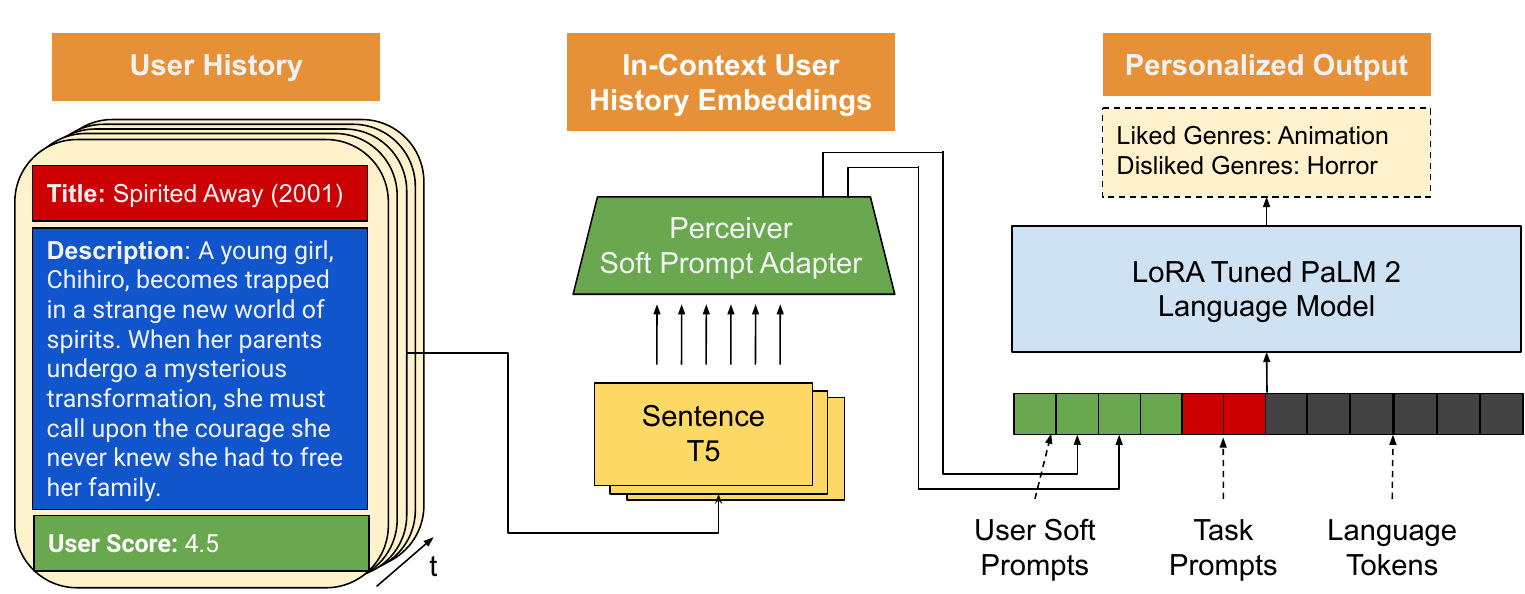}
    \caption{Overview of the PERSOMA architecture. Sentence T5 encodes natural language descriptions of the user's history and is then jointly resampled by a Perceiver adapter to create in-context user soft prompts. These soft prompts are then concatenated with task embeddings to prompt a PaLM 2 language model towards personalized output.}
    \label{fig:teaser}
\end{figure*}

Our method is grounded in the unified text-to-text approach proposed by T5 \cite{t5}, where tasks are conceptualized as text generation. Formally, for an input sequence of tokens $X$, the output sequence $Y$ is modelled probabilistically as $Pr_{\theta_{LM}}(Y|X)$ parameterized by the language model's learned weights $\theta_{LM}$. Prior research in text-to-text tasks can be divided into two predominant prompting methodologies: \textit{text-based prompting}, which integrates textual directives into the input \cite{DBLP:conf/acl/MishraKBH22, flant5, CoT}, and \textit{soft-prompting}, introducing a series of fixed trained tokens preceding the model's input tokens to capture information about a specific task \cite{lester2021power, prefixtuning}.

Conventionally, soft-prompting employs a static, task-specific soft-prompt to enhance parameter efficiency across various linguistic tasks, optimizing the likelihood $Pr_{\theta_{LM}}(Y|[T; X])$, where $T$ represents a fixed set of trainable task tokens (called task "soft prompts") and $X$ is the task input. Given a set of historical interactions $H_u$ described in natural language for user $u$, we further train a soft prompt adapter network to jointly \textit{compress and resample} these interactions into a set of personalized user=specific soft-prompt tokens $P_u$ where $|P_u| \leq |H_u|$. The generated soft-prompt tokens can then be utilized to optimize $Pr_{\theta_{LM}}(Y_u|[P_u; T; X])$, where $Y_u$ is a personalized text output conditioned on user $u$'s history. 

In practice, PERSOMA (depicted in Figure \ref{fig:teaser}) consists of three primary components: the history encoder, soft prompt adapter and resampler and a large language model decoder. We first encode each natural language user interaction $h_i \in H_u$ using a SentenceT5 text embedding model \cite{sentencet5}, our history encoder, creating $H'_u$. We then feed the set of $H'_u$ into our soft prompt adapter network to generate the set of personalized soft prompt tokens $P_u$. In our results, we experiment with architectures which jointly encode the sequence of $H'_u$ embeddings (Perceiver \cite{jaegle2021perceiver}, Transformer \cite{vaswani2017attention}) as well as those that encode embeddings individually (MLP projection). Finally, the combined soft prompt $[T; P_u]$ is fed into a frozen PaLM 2 Language Model to generate personalized responses to the given task parameterized by $T$. To ensure our work is comparable to the prior art, all mentions of PalM 2 refer to the smallest XXS version.

\subsection{MovieLens Personalized Genre Prediction Task}
\label{sec:task}
To train and evaluate PERSOMA, we utilize the personalized genre prediction task proposed by \cite{doddapaneni2024user} using the MovieLens dataset \cite{movielens}. We briefly describe it here for completeness. The MovieLens dataset contains the viewing habits of users and metadata about the movies they are watching. We formulate the genre prediction task on this dataset as follows: given a set of natural language titles $m^t$, descriptions $m^d$ and review scores $m^r$ for each movie $m \in H_u$ user $u$ watched, predict the set of genres the user likes $G_u^+$ and dislikes $G_u^-$. The task is suitable for evaluating personalization as the model must understand the user's preferences to predict personalized genre suggestions. We removed all films with less than 20 reviews to ensure that each movie was high quality and could be seen sufficiently across user sequences in the dataset. After pruning, the final dataset includes 14.4 M reviews from 127 K users spanning 8.2 K unique movies and 19 genres. 

To apply PERSOMA and other embedding-based personalization baselines on this task, we construct our set of item embeddings $H'_u$ by concatenating the history encoder embeddings for $m^t$, $m^d$ and $m^r$. For text-only baselines, we instead represent the input history by concatenating the direct text description of each movie, formatted as
$$U_{text} = \{[\texttt{Title:} m^t, \texttt{Description:} m^d, \texttt{Rating:} m^r] | ~\forall m \in H_u\}$$
Finally, to frame this task as a text-to-text generation task, we formulate the target genre prediction output $Y$ as 
$$Y = \texttt{Liked Genres:}~G_u^+ ~\texttt{Not Liked Genres:}~G_u^-$$ 
where a comma delimits each genre in $G_u^+$ and $G_u^-$ (ex: "Liked Genres: Action, Comedy. Not Liked Genres: Romance").   

Rather than use conventional text generation metrics such as BLEU and ROUGE, the predicted genre names are extracted from the generated string to evaluate class-weighted Recall, Precision, and F1. This ensures that we explicitly consider the model's performance on the target task, irrespective of the order in which the genres are predicted. Given that users can like or dislike each of the 19 genres, our evaluation metrics evaluate the problem as a 38-class multi-classification problem\footnote{The resulting multi-label distribution can be seen in the Appendix Table \ref{tab:genre_distribution}}. We divide the MovieLens dataset into sets of 114K/5K/5K users for training, validation and test respectively, following \cite{doddapaneni2024user}. 

\section{Results}
Using the MovieLens Personalized Genre Prediction Task, we evaluate our method against other state-of-the-art personalization baselines. We also conduct an extensive ablation study assessing the impact of various history sampling techniques and item input formats. Unless otherwise noted, we utilize the most recent $N$ movies the user watched to construct the user's history, where $N$ is either 5, 10 or 50 (denoted as "History $N$"). We also include a simple ``Counting'' baseline that selects the top three and bottom three watched genres from the user's history as the predicted liked and disliked genres, respectively.

In each experiment, we use 20 learned tokens as our task prompt $T$. We also use a batch size of 32 and a learning rate of 0.001, training for 300k steps with early stopping to optimize validation F1. When comparing various soft prompt adapter networks, we utilize a 3-layer MLP, 1-layer Transformer network, and 4-layer Perceiver network for each variant. For resampling with Perceiver, the output prompt size is resampled to 20 regardless of history size. We use PaLM2 XXS as the LLM in all our PERSOMA experiments.

\subsection{Effect of PERSOMA Soft Prompt Adapter Architecture}
\done
\begin{table*}
    \centering
    \caption{Performance of various adapter architectures and other personalization methods on the MovieLens Genre Prediction Task with Recency Sampling}
    \begin{tabular}{l|c|c|c|c|c|c|c|c|c}
    \toprule
                             & \multicolumn{3}{c}{History 5}
                             & \multicolumn{3}{c}{History 10} 
                             & \multicolumn{3}{c}{History 50} \\
        Encoder Architecture & F1 & Recall & Precision & F1 & Recall & Precision & F1 & Recall & Precision \\
    \midrule
        Counting & 0.177 & 0.223 & 0.246 & 0.181 & 0.251 & 0.249 & 0.168 & 0.248 & 0.271\\
    \midrule
        UEM Large & 0.215 & 0.290 & 0.281 & - & - & - & 0.381 & 0.399 & 0.400 \\
        UEM Base  & 0.252 & 0.297 & 0.275 & - & - & - & 0.396 & 0.405 & 0.407 \\
    \midrule 
        PERSOMA MLP         & 0.278 & 0.274 & 0.324 & 0.316 & 0.355 & 0.312 & 0.569 & 0.563 & 0.588 \\
        PERSOMA Perceiver   & 0.256 & 0.305 & 0.244 & 0.281 & 0.327 & 0.275 & 0.371 & 0.411 & 0.371 \\
        PERSOMA Transformer & 0.268 & 0.301 & 0.257 & 0.319 & 0.348 & 0.312 & 0.545 & 0.561 & 0.545 \\
    \midrule
        Fine-tuned PaLM 2 (XXS) Text Prompting & 0.285 & 0.316 & 0.279 & 0.335 & 0.360 & 0.327 & \multicolumn{3}{c}{Out of context length} \\
    \midrule\midrule
        PERSOMA MLP with Frozen LM          & 0.280 & 0.316 & 0.271 & 0.323 & 0.359 & 0.319 & 0.541 & 0.562 & 0.532 \\
        PERSOMA Perceiver with Frozen LM    & 0.260 & 0.311 & 0.265 & 0.297 & 0.331 & 0.292 & 0.339 & 0.378 & 0.360 \\  
        PERSOMA Transformer with Frozen LM  & 0.263 & 0.308 & 0.256 & 0.302 & 0.339 & 0.295 & 0.544 & 0.564 & 0.546 \\
    \midrule 
        Gemini 1.5 Pro Zero-Shot Text Prompting & 0.187 & 0.196 & 0.279 & 0.221 & 0.249 & 0.308 & 0.261 & 0.318 & 0.346 \\
    \bottomrule
    \end{tabular}
    \label{tab:persoma_encoder}
\end{table*}
A vital component of the PERSOMA architecture is how we process, resample, and represent the user's historical interactions for personalization. We evaluate PERSOMA's performance towards this capability by comparing it against various personalization methods when faced with a user history of the latest 5, 10, and 50 movies watched, respectively. We compare against end-to-end finetuned models UEM Base and UEM Large from \cite{doddapaneni2024user}, which use T5, and a PaLM 2 \cite{palm2} XXS model following the text input formulation described in section \ref{sec:task} as embedding-based and text prompt-based baselines respectively. We also include a zero-shot Gemini 1.5 Pro model baseline to compare against PERSOMA methods without end-to-end finetuning.

Our results, shown in Table \ref{tab:persoma_encoder}, comparing MLP, Perceiver, and Transformer variants of PERSOMA, demonstrate that both MLP and Transformer variants outperform UEM Base and UEM Large, particularly with larger history sizes. For example, PERSOMA MLP achieves a 0.19 higher F1 score than UEM Large when processing 50 items. Even with a frozen language model, PERSOMA MLP with Frozen LM still outperforms UEM Large by 0.16 F1 with the same history size.

Since PERSOMA MLP with Frozen LM trains far fewer parameters than UEM Large, our results indicate that finetuning the entire model may be unnecessary with a stronger base LLM (PaLM2 vs T5). Instead, training the adapter to create expressive user embeddings within the LLM language space is sufficient to create significant personalization gains, even at smaller history sizes. This result is further reinforced when resampling with a Perceiver model, unlocking further improvements in token efficiency.

\begin{table}
    \centering
    \caption{Average text token input length for Embedding-based versus Text-based methods. More tokens incur greater computational complexity.}
    \begin{tabular}{l|c|c|c}
        \toprule
        Method/History          &  5 &  10 &  50 \\
        \midrule
        Embedding-based     & 75        & 80       & 130 \\
        Embedding-based w Perceiver & 75      & 80      & 100  \\
        Text-based (99\%)          & 900       & 1600     & 16000 \\
        \midrule
        Relative Difference &\textbf{12x}       & \textbf{20x}      & \textbf{123x-160x} \\
        \bottomrule
    \end{tabular}
    \label{tab:input_length}
\end{table}

Next, we inspect the performance of PERSOMA against text-prompting baselines. Here, it is essential to note that text prompting requires significantly more computing, scaling according to text token length rather than having a single token per history item. This is best seen in Table \ref{tab:input_length}, where we note that text prompting would require $\approx$ 16 000 input tokens to model a user history of 50 items in MovieLens, whereas PERSOMA would only need 130, further reduced to 100 with Perceiver resampling. Further, it is worth noting that PaLM 2 is finetuned end-to-end toward the target task, tuning many more parameters than our Frozen variants.  

Despite utilizing several orders of magnitude less computing, the gap between PERSOMA and PaLM 2 text prompting is only a marginal 0.007 F1 and 0.019 F1 difference when using a history size of 5 and 10, respectively. Comparing PERSOMA Frozen and Gemini, we see that PERSOMA MLP achieves 0.308 higher F1 at a history length of 50. This demonstrates the effectiveness of using in-domain soft prompts for personalization over text prompts. 

Finally, we compare the performance of various PERSOMA soft prompt adapter architectures. Surprisingly, we find that PERSOMA MLP, a smaller and simpler architecture, outperforms PERSOMA Perceiver and PERSOMA Transformer. This is especially the case when utilizing a large history size of 50, where PERSOMA MLP outperforms PERSOMA Transformer and PERSOMA Perceiver by 0.024 F1 and 0.198 F1, respectively. 

There are a few possible reasons for this loss in optimal performance with sequential adapter models. First, it is essential to note that while MovieLens does provide an ordering of watched films, the order is self-reported by users on the platform via a survey \cite{movielens}. Given that ordering is not strict, architectures incorporating positional encodings, such as Transformer and Perceiver, may be better suited for tasks that heavily depend on interactions' temporal order. Second, the lack of performance of the Perceiver may be attributed to the reduced number of prompt tokens, which may not have sufficient expressivity as input to the LLM. However, we note that the Perceiver result resampled to a history length of 20 is better than MLP with a history length of 10.

\subsection{Parameter Efficient Training with LoRA}

\begin{table*}
    \centering
    \caption{Performance of various history sampling techniques and sizes. We evaluate PERSOMA MLP against PaLM 2 text prompting and naive counting.}
    \begin{tabular}{c|l|c|c|c|c|c|c|c|c|c}
    \toprule
    & 
                             & \multicolumn{3}{c}{History 5}
                             & \multicolumn{3}{c}{History 10} 
                             & \multicolumn{3}{c}{History 50} \\
        Sampling & Method & F1 & Recall & Precision & F1 & Recall & Precision & F1 & Recall & Precision \\
    \midrule
       \multirow{3}{*}{Recency}          & PERSOMA & 0.278 & 0.274 & 0.324 & 0.316 & 0.355 & 0.312 & 0.569 & 0.563 & 0.588 \\
                                         & PaLM 2 & 0.285 & 0.279 & 0.316 & 0.335 & 0.360 & 0.327 & \multicolumn{3}{c}{Out of context length} \\
                                         & Counting & 0.177 & 0.223 & 0.246 & 0.181 & 0.251 & 0.249 & 0.168 & 0.248 & 0.271 \\
    \midrule
       \multirow{3}{*}{Random}           & PERSOMA & 0.265 & 0.319 & 0.254 & 0.320 & 0.363 & 0.312 & 0.574 & 0.591 & 0.568 \\
                                         & PaLM 2 & 0.320 & 0.361 & 0.325 & 0.388 & 0.419 & 0.382 & \multicolumn{3}{c}{Out of context length} \\
                                         & Counting & 0.177 & 0.246 & 0.223 & 0.189 & 0.265 & 0.267 & 0.168 & 0.248 & 0.271\\
    \midrule
       \multirow{3}{*}{Long Tail}        & PERSOMA & 0.259 & 0.316 & 0.246 & 0.283 & 0.317 & 0.274 & 0.329 & 0.373 & 0.318 \\
                                         & PaLM 2 & 0.290 & 0.322 & 0.279 & 0.312 & 0.342 & 0.303 & \multicolumn{3}{c}{Out of context length} \\
                                         & Counting & 0.164 & 0.207 & 0.201 & 0.163 & 0.209 & 0.204 & 0.158 & 0.206 & 0.203 \\
    \midrule
       \multirow{3}{*}{Top-K Popularity} & PERSOMA & 0.294 & 0.358 & 0.325 & 0.351 & 0.389 & 0.355 & 0.593 & 0.613 & 0.588 \\
                                         & PaLM 2 & 0.324 & 0.374 & 0.325 & 0.393 & 0.433 & 0.396 & \multicolumn{3}{c}{Out of context length} \\
                                         & Counting & 0.183 & 0.254 & 0.281 & 0.189 & 0.267 & 0.314 & 0.166 & 0.249 & 0.296 \\
    \midrule
       \multirow{3}{*}{Genre Sample}     & PERSOMA & 0.271 & 0.336 & 0.259 & 0.313 & 0.367 & 0.321 & 0.587 & 0.607 & 0.580 \\
                                         & PaLM 2 & 0.313 & 0.349 & 0.313 & 0.370 & 0.388 & 0.360 & \multicolumn{3}{c}{Out of context length} \\
                                         & Counting & 0.189 & 0.257 & 0.243 & 0.187 & 0.267 & 0.237 & 0.166 & 0.251 & 0.269\\
    \bottomrule
    \end{tabular}
    \label{tab:sampling}
\end{table*}

\begin{table}
    \centering
    \caption{Performance of Parameter Efficient Techniques for Training PERSOMA MLP with a History Size of 50 Movies}
    \begin{tabular}{l|c|c|c}
        \toprule
        Method & F1 & Precision & Recall \\
        \midrule
        PERSOMA with LoRA LM    & 0.533 & 0.524 & 0.557 \\
        PERSOMA with Frozen LM  & 0.541 & 0.562 & 0.532 \\
        PERSOMA End-to-End  & 0.569 & 0.563 & 0.588 \\
        \bottomrule
    \end{tabular}
    \label{tab:lora}
\end{table}

To assess the effectiveness of parameter-efficient finetuning techniques for PERSOMA, we conducted experiments using Low-Rank Adaptation (LoRA) \cite{hu2021lora} and frozen language model weights, comparing them to a fully finetuned PERSOMA model. For LoRA experiments, we use rank four adaptation for both attention and transformer feedforward layers. 

As demonstrated in Table \ref{tab:lora}, when using recency sampling with a history size of 50, both LoRA and frozen PERSOMA variants achieved strong F1 scores of 0.533 and 0.541 respectively, closely matching the performance of the end-to-end finetuned model (0.569 F1). This indicates that parameter-efficient techniques can be effectively applied to PERSOMA, offering a promising avenue for reducing computational costs without sacrificing performance.

\subsection{Effect of History Sampling Strategies}
\done
Finally, inspired by the LAMP personalization benchmark \cite{lamp}, we extensively evaluate the impact of various methods sampling strategies on the performance of PERSOMA, PaLM 2 and Counting. For a given history size $H$, we evaluate the following sampling strategies: 
\begin{itemize}
    \item \textbf{Recency:} Select the latest $H$ watched movies
    \item \textbf{Random:} Uniformly sample $H$ watched movies 
    \item \textbf{Long Tail:} Uniformly sample $H$ watched movies that are below the global 90th quartile of popularity 
    \item \textbf{Top-K Popularity:} Select the top $H$ globally popular movies the user watched 
    \item \textbf{Genre Sample:} Sample $H$ movies the user watched according to the users genre density 
\end{itemize}

The results of each sampling strategy can be seen in Table \ref{tab:sampling}. We can see that PERSOMA performance improves significantly as the history size increases regardless of sampling strategy. PERSOMA achieved the highest performance with Top-K popularity, achieving 0.593 F1 and 0.294 F1 at history 50 and 5, respectively. Top-K sampling bested Recency sampling, which achieved 0.569 F1 and 0.278 F1 with a history of 50 and 5.  

One notable outlier sampling strategy was long tail sampling, achieving a lower F1 across PERSOMA and PaLM 2. This performance gap is likely due to many esoteric films not being in the popular zeitgeist and, therefore, not being well represented in the training data for the language model (unlike Top-K sampling, which features only popular movies). Besides long-tail sampling, all other strategies perform within $\pm$ 0.03 F1 of each other, demonstrating PERSOMA's robustness. 

\section{Conclusions \& Future Work}
\done
This paper introduces PERSOMA, an architecture designed to tackle the challenges of effectively modelling user history for personalization tasks. PERSOMA's core strength lies in its ability to compress and resample historical user interactions into informative in-context soft prompt tokens while employing parameter-efficient techniques for finetuning the language model.

Through extensive experimentation with various encoder architectures, sampling methods, and parameter-efficient techniques, we demonstrate the versatility and robustness of PERSOMA. Notably, PERSOMA matches the performance of text-based prompting even when restricted to the same history size and surpasses these baselines with longer histories, all while requiring significantly less computational power. This makes PERSOMA a valuable tool for practitioners seeking efficient natural language personalization without compromising performance.

We also believe that embedding representations in LLMs is a promising avenue for further work in personalized prompting.
We have investigated the effectiveness of our resampling strategies with Perceiver on the MovieLens dataset, which contains user sequences of O(100) history items. However, real-world production datasets often contain much more extensive interaction histories. Exploring the benefits of resampling with such datasets would be valuable. 

Additionally, incorporating sparse first-stage retrievers to pre-filter history items for PERSOMA could prove beneficial, mainly when dealing with histories exceeding 500 interactions. Our sampling experiments, specifically the positive impact of task-targeted sampling (Top-K and Genre Sampling), highlight the potential of this approach. Combining these techniques with Perceiver offers a promising way to efficiently represent long user journeys.

Finally, while we employed PaLM 2 as our large language model decoder, future studies could examine the performance of various LLM decoders of different sizes, such as replacing PaLM 2 with T5 small \cite{t5} Phi-2 \cite{li2023textbooks}, or MiniLM \cite{wang2020minilm}. Such research could shed light on the applicability of PERSOMA in low-latency and resource-constrained production environments.

\begin{acks}
The authors would like to thank Santiago Ontanon, who graciously offered expert advice and feedback on initial versions of this work. 
\end{acks}

\bibliographystyle{ACM-Reference-Format}
\bibliography{base}

\appendix
\section{Label distribution}
\begin{table}
    \centering
    \caption{Label Distribution of the Movie Lens Genre Prediction Task. In this task, training examples consist of sets of ``liked'' and ``disliked'' genres}
    \begin{tabular}{l|c|c|c}
    \toprule
        Genre & Like  & Dislike & Total  \\
    \midrule
        Drama & 41 404 & 1 747    & 43 151 \\
        Crime & 30 243 & 2 249    & 32 489 \\
        War   & 30 708 & 1 204    & 31 912 \\
        Romance & 23 830 & 3 879  & 27 709 \\
        Thriller & 20 239 & 3 711 & 23 950 \\
        Comedy & 16 972 & 5 369 & 22 341 \\
        Mystery & 20 106 & 1 872 & 21 978 \\
        Adventure & 14 589 & 4 894 & 19 483 \\
        Action & 12 665 & 5 649 & 18 314 \\
        Fantasy & 12 653 & 4 154 & 16 807 \\
        Animation & 14 389 & 2 293 & 16 682 \\
        Children & 10 870  & 5 195 & 16 065 \\
        IMAX & 9 277 & 1 892 & 11 169 \\
        Horror & 5 972 & 5 078 & 11 050 \\
        Musical & 7 510 & 2 133 & 9 643 \\
        Western & 4 689 & 757 & 5 446 \\
        Documentary & 3 353 & 178 & 3 531 \\
        Sci-Fi & 1 388 & 498 & 1 886 \\
        Film-Noir & 681 & 16 & 697 \\
        \midrule
        All & 281 538 & 52 765 & 334 303 \\
    \bottomrule
    \end{tabular}
    \label{tab:genre_distribution}
\end{table}

\section{Effect of Item Input Format}
\begin{table*}
    \centering
    \caption{Ablation study of various item input formats on PERSOMA and PaLM 2 text prompting}
    \begin{tabular}{c|l|c|c|c|c|c|c|c|c|c}
    \toprule
    & 
                             & \multicolumn{3}{c}{History 5}
                             & \multicolumn{3}{c}{History 10} 
                             & \multicolumn{3}{c}{History 50} \\
        Input Format & Method & F1 & Recall & Precision & F1 & Recall & Precision & F1 & Recall & Precision \\
    \midrule
       \multirow{2}{*}{Complete}         & PERSOMA & 0.250 & 0.321 & 0.246 & 0.315 & 0.355 & 0.312 & 0.563 & 0.576 & 0.560 \\
                                         & PaLM 2  & 0.285 & 0.279 & 0.316 & 0.335 & 0.360 & 0.327 & \multicolumn{3}{c}{Out of context length} \\
    \midrule
       \multirow{2}{*}{Title + Score}    & PERSOMA & 0.303 & 0.330 & 0.302 & 0.355 & 0.376 & 0.353 & 0.558 & 0.598 & 0.591 \\
                                         & PaLM 2  & 0.280 & 0.294 & 0.272 & 0.339 & 0.354 & 0.328 & \multicolumn{3}{c}{Out of context length} \\
    \midrule
       \multirow{2}{*}{Desc. + Score}    & PERSOMA & 0.275 & 0.321 & 0.286 & 0.326 & 0.365 & 0.328 & 0.523 & 0.536 & 0.521 \\
                                         & PaLM 2  & 0.307 & 0.343 & 0.299 & 0.373 & 0.392 & 0.377 & \multicolumn{3}{c}{Out of context length} \\
    \bottomrule
    \end{tabular}
    \label{tab:no_desc_title}
\end{table*}
A key aspect of PERSOMA is its utilization of semantic content within historical items. To assess the sensitivity of our architecture and the personalization task to different representations, we conducted an ablation study examining the impact of removing either the movie title or description (Table 5).

Our findings reveal that, for shorter history lengths, PERSOMA performs better using only the movie title than the complete description. Interestingly, omitting the title affects PERSOMA more significantly than omitting the description, while the opposite effect is observed with text prompting using PaLM 2. We hypothesize that this is because movie titles often succinctly capture the essence of a film, unlike a detailed description. Consequently, it is more challenging to learn the compression from a descriptive embedding into a limited number of tokens (5 or 10).

In scenarios with short history inputs, text prompting techniques may be preferable. Alternatively, increasing the size of output tokens independently of the input history length (i.e., resampling to a larger token length) could be considered. We leave this exploration for future work.

\end{document}